\documentclass[runningheads]{llncs}
\usepackage{graphicx}
\usepackage{amsmath,amssymb} 
\usepackage{color}
\usepackage[width=122mm,left=12mm,paperwidth=146mm,height=193mm,top=12mm,paperheight=217mm]{geometry}

\graphicspath{{./figure/}}

\begin{document}

\pagestyle{headings}
\mainmatter

\title{Open-set Person Re-identification} 

\titlerunning{Open-set Person Re-identification}

\authorrunning{Shengcai Liao et al.}

\author{Shengcai Liao, Zhipeng Mo, Jianqing Zhu, Yang Hu, and Stan Z. Li}
\institute{Center for Biometrics and Security Research\\
National Laboratory of Pattern Recognition\\
Institute of Automation, Chinese Academy of Sciences\\
95 Zhongguancun East Road, Beijing 100190, China\\
{\tt\small \{scliao,jianqing.zhu,yhu,szli\}@nlpr.ia.ac.cn}}

\maketitle

\begin{abstract}
Person re-identification is becoming a hot research for developing both machine learning algorithms and video surveillance applications. The task of person re-identification is to determine which person in a gallery has the same identity to a probe image. This task basically assumes that the subject of the probe image belongs to the gallery, that is, the gallery contains this person. However, in practical applications such as searching a suspect in a video, this assumption is usually not true. In this paper, we consider the open-set person re-identification problem, which includes two sub-tasks, detection and identification. The detection sub-task is to determine the presence of the probe subject in the gallery, and the identification sub-task is to determine which person in the gallery has the same identity as the accepted probe. We present a database collected from a video surveillance setting of 6 cameras, with 200 persons and 7,413 images segmented. Based on this database, we develop a benchmark protocol for evaluating the performance under the open-set person re-identification scenario. Several popular metric learning algorithms for person re-identification have been evaluated as baselines. From the baseline performance, we observe that the open-set person re-identification problem is still largely unresolved, thus further attention and effort is needed.

\keywords{person re-identification, open-set identification, database and benchmark}
\end{abstract}

\section{Introduction}
Video surveillance is an important technique for both security and forensic applications. However, since the volume of surveillance videos have become larger and larger today, searching desired objects from massive surveillance videos turns out to be very difficult. Person re-identification is one of the useful technique to facilitate this head-scratching search. It is becoming a hot research for developing both machine learning algorithms and video surveillance applications. Many approaches have been proposed for person re-identification \cite{Vezzani-CSUR-2013,gong2014person}, which greatly advance this field.

The task of person re-identification is to determine which person in a gallery has the same identity to a probe image. This task basically assumes that the subject of the probe image belongs to the gallery, that is, the gallery contains this person. This scenario is called the closed-set identification \cite{Phillips-HFR2}. Under this assumption, the evaluation of person re-identification algorithms is done by performing the top-k retrieval and determining whether the true match appears in the result. Furthermore, the Cumulative Matching Characteristic (CMC) curve is used to illustrate the identification accuracy versus the number of samples retrieved. Many databases have been created to test person re-identification algorithms under the closed-set scenario, as summarized in Table \ref{tab:databases}.

A typical application of person re-identification in forensic video analysis is to search a suspect in a video or a large set of videos. In this application, the probe image contains a suspect (e.g. from a shot of a surveillance video). The task is to determine whether this suspect appears in a certain video, and if it does, when and where this person appears. In this application, there is no assumption that the person in the probe image is present in the searching video. Instead, the system needs to answer whether the query person appears in the searching video, and if it does, which person in the video corresponds to the query person. Without the ability of determining the presence of a suspect in a video, one can also apply available person re-identification algorithms to retrieve the top-k candidates, and browse the results to determine whether the suspect is found. However, considering that there may be a large set of videos to be searched, it still requires large human labors to check top-k candidates from every video. Instead, if a strong evidence indicates that the searching video does not contain the suspect, the video can be automatically discarded to save time and labor.

The open-set person re-identification problem studied in this paper is oriented for the above forensic application of searching a suspect in a recorded surveillance video or a large set of videos (treated as the gallery). In this application, two requirements for person re-identification algorithms are generally needed: i) algorithm should be trained with an independent data set, because it is not practical to learn a classifier or learn person models every time a video comes; and ii) algorithm should be able to answer whether the probe person is contained in the video or not. This problem is similar as the open-set face identification problem proposed in \cite{Phillips-HFR2,Liao-PAMI-2013,Liao-IJCB-14}. Formally, open-set person re-identification includes two sub-tasks, detection and identification. The detection sub-task is to determine the presence of the probe subject in the gallery, and the identification sub-task is to determine which person in the gallery has the same identity as the accepted probe.

Accordingly, in this paper, we present a database from a video surveillance setting of 6 cameras, with 200 persons and 7,413 images segmented. Based on this database, we develop a benchmark protocol for evaluating the performance under the open-set person re-identification scenario. Basically, the proposed open-set person re-identification benchmark protocol involves four subsets, namely, training set, gallery set, genuine probe set, and impostor probe set, where test set = gallery set + probe set, and probe set = genuine probe set + impostor probe set (see Fig. \ref{fig:OPeRID} for an illustration). With this setting we introduce two kinds of open-set problems: i) persons of the test set are not known from the training set; and ii) persons of the impostor probe set are not known from the gallery set. We require learning a general camera independent person re-identification model from the training set, so as to test the algorithm's generalization ability to unseen persons and unknown camera views.

Several popular metric learning algorithms for person re-identification have been evaluated as baselines in this paper. However, from the baseline performance, we observe that the open-set person re-identification problem is still largely unresolved. Therefore, further effort should be spent to tackle this problem. We make the collected database (called Open-set P\textit{e}rson Re-IDentification database, OPeRID v1.0), as well as the benchmark tool and the extracted features publicly available in a project website\footnote{http://www.cbsr.ia.ac.cn/users/scliao/projects/operidv1/} to promote further development along this direction.

The remaining paper is organized as follows. In Section \ref{sec:review}, we briefly review the related work. In Section \ref{sec:database}, we describe the collection of the database. In Section \ref{sec:benchmark}, we introduce the benchmark protocol for the open-set person re-identification. We introduce several metric learning algorithms as the baseline methods in Section \ref{sec:method}, and present the baseline performance in Section \ref{sec:baseline}. Finally, we conclude the paper in Section \ref{sec:summary}.

\section{Related Work}\label{sec:review}
Recently, more and more benchmark datasets for person re-identification are available. These datasets contain images of persons with variations in resolutions, lightings, poses, occlusion, and background in the same camera view and across camera views and thus make the person re-identification task very challenging. We summarize some popular person re-identification datasets in Table \ref{tab:databases}. Among them, the VIPeR dataset \cite{gray2007evaluating} is one of the earliest single-shot datasets in this field, and it is the most widely used benchmark dataset so far. The ETHZ dataset \cite{ess2007depth,schwartz2009learning} contains images captured from a single moving cameras in a street scene, therefore the main challenge is occlusions rather than viewpoint variations. The i-LIDS MCTS dataset \cite{nilski2008evaluating,zheng2011person} was collected from an airport arrival hall, with some persons having images from the same camera view, while others having images from different non-overlapping camera views. The CAVIAR4REID dataset \cite{cheng2011custom} contains 1,220 images of 72 pedestrians with large variation in resolutions, but only 50 of the 72 pedestrians appear in two camera views. The PRID2011 dataset \cite{hirzer11a} contains 931 persons from two camera views, with 200 persons appearing in both camera views. The GRID dataset \cite{loy2009multi} was captured from 8 disjoint camera views in a underground station. It contains 250 pedestrian image pairs, with each pair containing two images of the same person from different camera views. Besides, there are 775 additional images that do not belong to the 250 persons, which can be used to enlarge the gallery. The 3DPES dataset \cite{baltieri20113dpes} contains 200 persons from 8 different surveillance cameras. An advantage of this dataset is that some persons can be seen from 3 different camera views. The CUHK02 dataset \cite{Li-CVPR-2013-CUHK02} contains 1,816 persons captured from 5 pairs of cameras. It has the largest number of persons so far. More details and characteristics of the above mentioned datasets are listed in Table \ref{tab:databases}.
\begin{table}[!hbt]
    \normalsize
    \centering
    \caption{Some popular datasets used for person re-identification}\label{tab:databases}
    \begin{center}
        \begin{tabular}{|c|c|c|c|c|c|}
            \hline
             \textbf{Dataset} & \textbf{$\#$Cams} & \textbf{$\#$Person} & \textbf{$\#$Imgs} & \textbf{$\#$Views} & \textbf{Scenario}\\
            \cline{1 - 5}
            \hline
            VIPeR \cite{gray2007evaluating} & 2 & 632 & 1,264 & 2 &Outdoor \\
            \hline
            i-LIDS \cite{nilski2008evaluating,zheng2011person}& 5 & 119 & 476 & 2 &Indoor, Airport\\
            \hline
            ETHZ \cite{ess2007depth,schwartz2009learning} & 3 & 146 & 8,555 & 2 & Outdoor,City Street\\
            \hline
            CAVIAR4REID \cite{cheng2011custom} & 2 & 72 & 1,220 & 2 & Indoor, Shopping Center\\
            \hline
            3DPeS \cite{baltieri20113dpes} & 8 & 200 & 1,012 & 3 & Outdoor, Campus\\
            \hline
            PRID2011 \cite{hirzer11a} & 2 & 200 & 1,134 & 2 & Outdoor\\
            \hline
            CUHK02 \cite{Li-CVPR-2013-CUHK02} & 5 pairs & 1,816 & 7,264 & 2 & Outdoor, Campus\\
            \hline
            GRID \cite{loy2009multi} & 8 & 250 & 1,275 & 2 & Indoor, Underground\\
            \hline
        \end{tabular}
    \end{center}
    \scriptsize Note: Column three for PRID2011 and GRID contains only persons appearing in both views. $\#$Views denotes the maximum number of camera views for the same person.
\end{table}

The emergence of all these benchmark datasets have promoted the development of person re-identification to a large degree. However, they also have certain drawbacks that need to be addressed \cite{gong2014person}. Here, we concern about the number of persons, samples, and the number of camera views that a person has. Most existing databases do not have enough data to support the four-subset division for open-set person re-identification as aforementioned. In fact, a small training set will not be adequate for algorithm learning; a small gallery set will make the identification too easy; and a small impostor probe set will lead to a limited FAR range. Furthermore, existing databases have limited camera views, which usually lead to a single-view probe set. To the best of our knowledge, for a specified person, no more than two cameras are related in most datasets (3DPES has some persons seen from 3 different cameras), although the maximum number of cameras is already eight. Therefore, we collected a new database to better support the open-set person re-identification benchmark, with a sufficient training set, larger gallery and probe sets, and diverse camera views for the probe set to challenge the matching. Note that having five camera views of a person does not make the problem easier. On the contrary, this makes the problem more difficult because we put only one view in the gallery and keep all the other views independently (not associated a priori) in the probe set to make it diverse for open-set re-identification.

In the literature, several papers have addressed some similar problems of open-set person re-identification \cite{Bauml-AVSS-11,Bedagkar-Gala-ICCVW-11,Eisenbach-AVSS-12,Layne-VSReID-14}. In \cite{Bauml-AVSS-11,Bedagkar-Gala-ICCVW-11,Eisenbach-AVSS-12}, the gallery set was used as the training set, and classifiers were learned to model the known persons. Then, using the learned classifiers, a probe image can be classified as a known person or an outlier. This is an open-set person re-identification issue, however, they did not consider to learn a general model independent of the training set, so the learned models cannot be applied to unknown galleries. Li et al. \cite{Li-TCSVT-14} proposed an open-set person verification scenario, where only binary classification was considered, namely, whether two test images belong to the same person or not. Recently, Layne et al. \cite{Layne-VSReID-14} proposed an open-world person re-identification scenario, where a flying drone was used to patrol an area, collect data, and re-identify persons. Three tasks were proposed in \cite{Layne-VSReID-14}: watch-list, intra-flight re-identification, and inter-flight re-identification. While the intra-flight and inter-flight re-identification tasks are more like a continuous single-camera or multi-camera tracking problem, the watch-list task (identify a certain watch-list person from the patrol video) is more interesting, and closely related to the open-set forensic search problem addressed in this paper. For evaluation, the conventional information-retrieval style metrics are used in \cite{Layne-VSReID-14}, namely, the rank of the true matches, and the precision-recall curves. However, the rank metric is not integrated with the precision metric; therefore, the two independent metrics make the performance comparison not intuitive. In contrast, we utilize the well established open-set face recognition performance evaluation method \cite{Phillips-HFR2,Liao-PAMI-2013,Liao-IJCB-14} for the open-set person re-identification evaluation in this paper, where the rank of the true matches and the false accept rate are integrated to form a unified metric.


\section{Database Collection}\label{sec:database}

\begin{figure}
    \center
    \includegraphics[width=80mm] {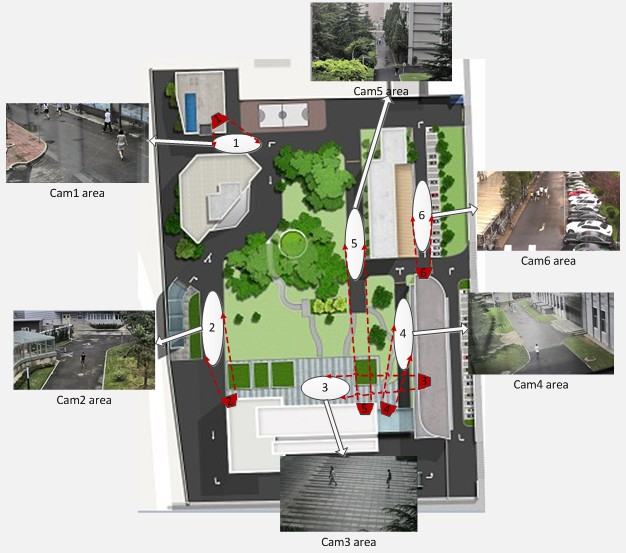}\\
    \caption{Illustration of the camera setup. The white ellipse areas indicate the FOV of a single camera, and the red trapezoidal blocks denote the positions where we locate the cameras. Six snapshots of the video frames are shown near the corresponding camera views to illustrate the captured scenes.}\label{fig:camera-layout}
\end{figure}

Our database collection was performed in a real outdoor surveillance scenario (institute campus) covered by a network of six distributed cameras, with the camera network setting illustrated in Fig. \ref{fig:camera-layout}, where cameras have non-overlapping field of views (FOVs). The FOVs selected correspond to common pedestrian walkways, with different camera viewpoints and scene illumination conditions. Therefore, a person's appearance vary from FOV to FOV when the person walks through the monitored areas. We utilized three kinds of cameras to collect the video. For Cam1, Cam2, Cam4 and Cam5, videos were collected by the HIKVISION DS-2CD864FWD-E IP camera. For Cam3, we utilized the SAMSUNG SNB-6004P IP camera. The SAMSUNG SNB-7000P IP camera was used for Cam6. We set the image resolution to $1280\times720$, and the frame rate is 25fps for all the six cameras. In order to capture clear images of pedestrians with more details,  we set the camera focal length to the largest and use auto iris. Nevertheless, underexposure and occlusions may exist as we capture the videos at busy time in a cloudy day. Finally, we obtain a 125-minute video from each camera.

We further segmented two sessions of video clips from the collected videos, with one session intended for training, and the other one for test. The two sessions are about 25 minutes apart. Each session contains six video clips of about 20 minutes, with each video corresponding to one camera view. From the 12 video clips, we segmented 7413 images of 200 persons using an interactive labeling software developed by \cite{Vondrick-IJCV-2013}. The statistics of these images are summarized in Table \ref{tab:database}. The resolutions of these segmented images varying from $16\times45$ to $135\times360$.
\begin{table}
\centering
\caption{Number of persons and images in each camera view of each session.}\label{tab:database}
\begin{tabular}{|c|c|c|c|c|}
  \hline
  &\multicolumn{2}{|c|}{Training session} & \multicolumn{2}{|c|}{Test session} \\
  \cline{2-5}
           &\#persons&\#images&\#persons&\#images \\
           \hline
  Camera 1 & 41     & 468   & 58    & 1,026 \\
  Camera 2 & 8      & 166   & 13    & 525 \\
  Camera 3 & 52     & 697   & 92    & 1,392 \\
  Camera 4 & 59     & 1,000 & 74    & 1,033 \\
  Camera 5 & 34     & 283   & 39    & 329 \\
  Camera 6 & 23     & 272   & 17    & 222 \\
  Total    & 81     & 2,886 & 119   & 4,527 \\
  \hline
\end{tabular}
\end{table}

One advantage of the collected database is that a person may have up to 5 camera views. This is counted in Table \ref{tab:camera-views}. Note that each of the 200 person has at least 2 associated camera views.
\begin{table}
\centering
\caption{Number of persons that have multi camera views.}\label{tab:camera-views}
\begin{tabular}{|c|c|c|}
  \hline
  &{Training session} & {Test session} \\
  \hline
  2 camera views & 33 & 70\\
  \hline
  3 camera views & 43 & 43\\
  \hline
  4 camera views & 3 & 6\\
  \hline
  5 camera views & 2 & 0\\
  \hline
\end{tabular}
\end{table}

\begin{figure}
\centering
Cam 1:~~ \hspace{8mm}
\hspace{8mm}
\includegraphics[width=8mm]{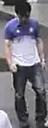}
\includegraphics[width=8mm]{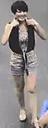}
\includegraphics[width=8mm]{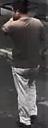}
\includegraphics[width=8mm]{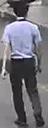}\\
\vspace{1mm}
Cam 3:~~ \includegraphics[width=8mm]{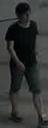}
\includegraphics[width=8mm]{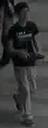}
\hspace{8mm}
\hspace{8mm}
\includegraphics[width=8mm]{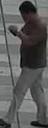}
\hspace{8mm}~\\
\vspace{1mm}
Cam 4:~~ \includegraphics[width=8mm]{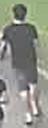}
\includegraphics[width=8mm]{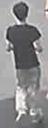}
\includegraphics[width=8mm]{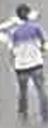}
\includegraphics[width=8mm]{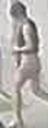}
\hspace{8mm}
\includegraphics[width=8mm]{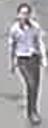}\\
\vspace{1mm}
Cam 5:~~ \hspace{8mm}
\hspace{8mm}
\includegraphics[width=8mm]{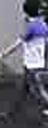}
\includegraphics[width=8mm]{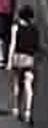}
\hspace{8mm}
\includegraphics[width=8mm]{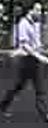}
\caption{Example images from the collected database. Images in the same column represent the same person. The four rows correspond to the camera views of 1, 3, 4, and 5, respectively.}
\label{fig:samples}
\end{figure}

All segmented images are scaled to $128\times48$ pixels. Fig. \ref{fig:samples} shows some example pairs of images from the collected database. It can be observed that there is a large variation in the observed color, and there are also lighting changes and viewpoint changes that challenge the matching of persons across cameras. Besides, though all videos we recoded are high definition videos in resolution, there still exist low resolution and blur pedestrian images due to long surveillance distance.

\section{Benchmark Protocol}\label{sec:benchmark}
\subsection{Experimental Setting}
With the collected database, we design the open-set re-identification experiments as 10 random trials including training and test, and each trial involves about half of the data. The detailed procedure is described as follows.

\subsubsection{Training}
Training is performed with the data of the training session. For each trial, half of the persons under each camera are randomly selected. If a person in a camera is selected, all images of that person within the camera can be used for training. Both within-camera pairs and between-camera pairs of samples can be used for learning the similarity metric. However, it is not allowed to learn a specific camera-to-camera transition model, and apply this model to the same camera pairs for testing. Though our test data has the same camera setting as the training data, we still require learning a general camera independent person re-identification model, so as to test the algorithm's generalization ability to possibly unknown camera pairs. This is also preferred in practical applications, because it is not easy to re-train the algorithm every time when dealing with new camera pairs. Note that if learning based dimension reduction method (e.g. PCA) is applied, it should be trained in each of the 10 trials for fair comparison.

\subsubsection{Test}
Test is performed by applying the learned model on the test session data. For each trial, also half of the persons under each camera are randomly selected. The difference to the training data selection is that, if a person in a camera is selected, the same person in all other cameras is also selected for test. If a person is selected, all images of that person are used for test. For each trial, the learned model of the corresponding trial is applied to the randomly selected test data, and the score of each between-camera sample pair is calculated for performance evaluation.

\subsubsection{Parameter Selection}
If one algorithm needs to determine the optimal hyper parameters, a cross validation procedure can be done with the training session data. Alternatively, the 10 random trials mentioned in the training procedure can also be utilized, and use the remaining half data of each trial for validation. However, any parameter tuning with the test session data is not allowed. This is to prevent hyper parameter optimization with the test data. A good algorithm should have a good generalization ability of both models and hyper parameters on unseen test data.

\subsection{Performance Evaluation}
In the performance evaluation, the open-set person re-identification scenario is considered. As indicated in~\cite{Phillips-HFR2}, the open-set identification task is a more general scenario, with the closed-set identification being its special case. Two sub-tasks, detection and identification, are involved in the open-set identification process. In the detection sub-task, the system decides whether the person of the probe image presents in the gallery or not. In the identification sub-task, the system reports the identity of the accepted probe. Therefore, the task of open-set identification is to determine the identity of the probe or to reject the probe.

The performance evaluation of the open-set person re-identification task involves three sets of images. The first set is the gallery set $G$, which contains person images collected by the system (e.g. pedestrians detected and segmented from a video). The other two are probe sets $P_G$ and ${P_N}$. While ${P_G}$ consists of persons in the gallery set ${G}$ but with different images, ${P_N}$ includes persons that are not present in ${G}$. Two performance measures, the detection and identification rate ($DIR$), and the false accept rate ($FAR$), are calculated for evaluation~\cite{Phillips-HFR2,Liao-PAMI-2013,Liao-IJCB-14}. Let $id(g,p)$ be an indicator whether $g$ and $p$ belong to the same identity, that is,
\begin{equation}
id(g,p) = \left\{
\begin{split}
&1,~g~\text{and}~p~\mbox{belong to the same identity},\\
&0,~\text{otherwise}.
\end{split}
\right.
\end{equation}
Let $s(\cdot,\cdot)$ be the similarity score function. Let
\begin{equation}
g^*=arg\max_{g\in{G}, id(g,p)=1} s(g,p),
\end{equation}
that is, $g^*$ is the gallery image that has the same identity as the probe image $p$ and reaches the maximum score among all gallery images of the same identity. Furthermore, let $rank(p)$ denote the rank order of $s(g^*,p)$ among matching scores between $p$ and all gallery images. That is, $rank(p)=k$ means that $s(g^*,p)$ is the $k^{th}$ largest similarity score. Then, the $DIR$ and $FAR$ measures are formulated as
\begin{equation}\label{equ:dir}
DIR(\tau, k) = \frac{|\{p|p\in{P_G},~rank(p)\le k,~s(g^*,p)\ge\tau\}|}{|{P_G}|},
\end{equation}
\begin{equation}\label{equ:far}
FAR(\tau) = \frac{|\{p|p\in{P_N},~\text{and}~\max_{g\in G} s(g,p)\ge\tau\}|}{|{P_N}|},
\end{equation}
where $\tau$ is the decision threshold, and $|A|$ calculates the number of elements in the set $A$.

Given a rank level $k$, by changing the decision threshold $\tau$, a Receiver Operating Characteristic (ROC) curve can be drawn by plotting $DIR$ versus $FAR$. Given an $FAR$ level, the operating threshold $\tau$ can be determined by Eq. (\ref{equ:far}), and a CMC curve of $DIR$ versus the rank $k$ can also be plotted. Note that when FAR=100$\%$, the corresponding $DIR$ is the traditional identification rate of the closed-set identification task, with the gallery set ${G}$ and the probe set ${P_G}$.

In the evaluation of the open-set person re-identification, for each trial, we construct the three sets $G$, $P_G$ and $P_N$ using all the randomly selected test images. First, for a camera $c$, a gallery set $G_c$ is constructed using all the selected test images from that camera. Second, images from other cameras with persons belonging to $G_c$ are used to construct a probe set $P_{G_c}$, and all other images form the other probe set $P_{N_c}$. The three sets are used to evaluate the open-set person re-identification performance measures $DIR_c$ and $FAR_c$ according to Eqs. (\ref{equ:dir}) and (\ref{equ:far}). Then, each camera is used to construct a gallery set in tune, and all performance measures $DIR_c$ and $FAR_c$ are averaged, respectively, to get the mean performance of the trial. Finally, performance measures of all the 10 random trials are averaged.

Furthermore, we also compute the standard deviation over the 10 trials of the performance measures, and we adopt the fused performance measure $\mu-\sigma$ as proposed in \cite{Liao-IJCB-14}, which considers both the mean and standard deviation statistics for performance evaluation. The purpose of the fused measure is to enforce comparison of the standard deviation. This metric requires algorithms to be stable in all random trials.

\begin{figure}
\centering
\includegraphics[width=80mm]{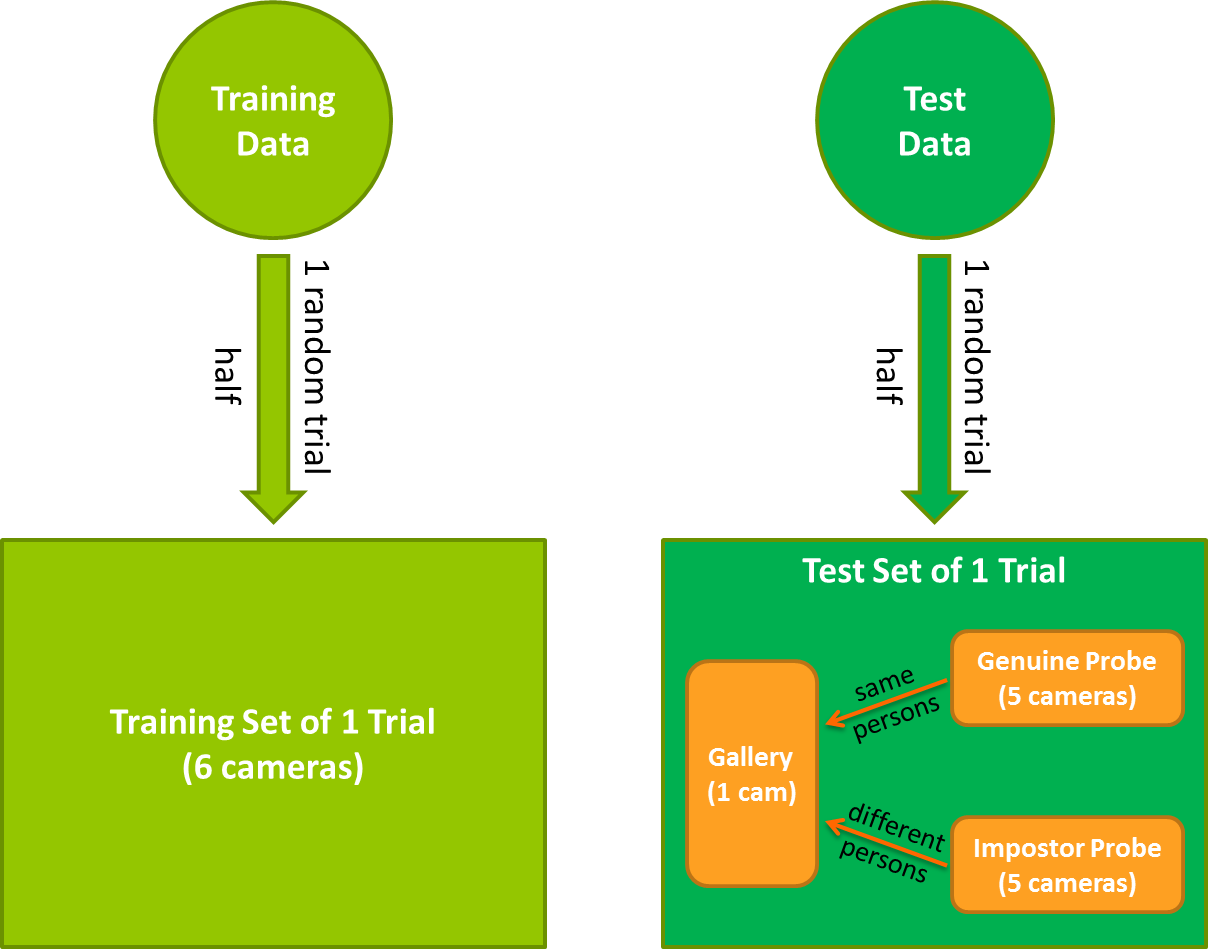}
\caption{Illustration of the OPeRID performance evaluation. For each trial, half of the training data is randomly sampled to learn a general distance or similarity metric, and half of the test data is also randomly sampled for performance evaluation. In addition, the randomly sampled test data is divided into the gallery set $G$ with a certain camera, the genuine probe set $P_G$ containing the same persons as in $G$ but with the other five cameras, and the impostor probe set $P_N$ containing different persons from $G$ who appear in the other five cameras. This test set partition is repeated for six times where each time a gallery set is constructed with one of the six cameras in tune, and the performance is averaged. The above procedure is repeated for 10 trials, and the mean and standard deviation of the performance is calculated.}
\label{fig:OPeRID}
\end{figure}

The above evaluation procedure is illustrated in Fig. \ref{fig:OPeRID}.

\section{Baseline Methods}\label{sec:method}
\subsection{Feature Representation}
For feature representation, we utilize a sliding window based histogram representation. Three scales of subwindows, $16\times16$, $32\times32$, and $48\times48$, have been used. The sliding step is set to be 8 pixels. In a $128\times48$ image, this generates 125 subwindows in total.

In each subwindow, we extract both color and texture features. For color features, the HSV color space is adopted. The H channel is quantized as 8 bins, while the S and V channels are quantized as 3 bins. Then, a joint histogram of $8\times8\times3$ bins is extracted in each subwindow. For the texture feature, we adopt the Multi-scale Block based Local Binary Patterns (MB-LBP) \cite{Liao-ICB-07}. Three scales of MB-LBP, $3\times3$, $7\times7$, and $11\times11$, have been used. A rotation invariant encoding \cite{Ojala-PAMI-02} is applied, resulting a 30-bin histogram in a subwindow for the three scales of MB-LBP. The color and texture histograms of all subwindows are concatenated to build the final descriptor. To suppress the influence of very large feature values, we truncated each histogram bin values to be no larger than 255, and then made a square-root transform. We make all the extracted features from our collected database publicly available in the project website, so that future classifier or metric learning algorithms can be applied on the same feature set for evaluation.

The dimensions of our feature descriptor is 12,750. We apply the Principle Component Analysis (PCA) approach in each trial to reduce the dimensions to 100 for further metric learning.

\subsection{Metric Learning}
We used a software package provided by \cite{kostinger2012large} to evaluate several metric learning algorithms, including Identity (i.e. Euclidean distance), MAHAL (i.e. Mahalanobis distance learned from positive sample pairs), LMNN \cite{weinberger2006distance}, ITML \cite{davis2007information} and KISSME \cite{kostinger2012large}. The LMNN algorithm \cite{weinberger2006distance} aims at learning a Mahalanobis distance metric for improving the k-nearest neighbor (kNN) classification, which can be solved by semidefinite programming. The ITML approach \cite{davis2007information} considers minimizing the differential relative entropy between two multivariate Gaussians for learning the Mahalanobis distance function. The KISSME algorithm \cite{kostinger2012large} applies the log likelihood ratio test of two Gaussian distributions for metric learning, and so a simplified solution can be derived. Besides, we also utilized the LADF algorithm (Locally-Adaptive Decision Functions) \cite{li2013learning} for evaluation. This algorithm can be viewed as a joint model of a distance metric and a locally adapted thresholding rule. Furthermore, we implemented a linear regression based discriminant analysis method introduced in \cite{Hastie-Book-2009}. This method learns a discriminant subspace via linear regression of class labels. Different from \cite{Hastie-Book-2009}, we applied the ridge regression method instead of the least square regression for robust learning. We call the resulting algorithm Ridge Regression based Discriminant Analysis (RRDA). We further used the Cosine similarity metric on the derived subspace to measure the similarity between two pedestrian images.

\section{Baseline Performance}\label{sec:baseline}
\begin{figure}
\centering
\includegraphics[width=80mm]{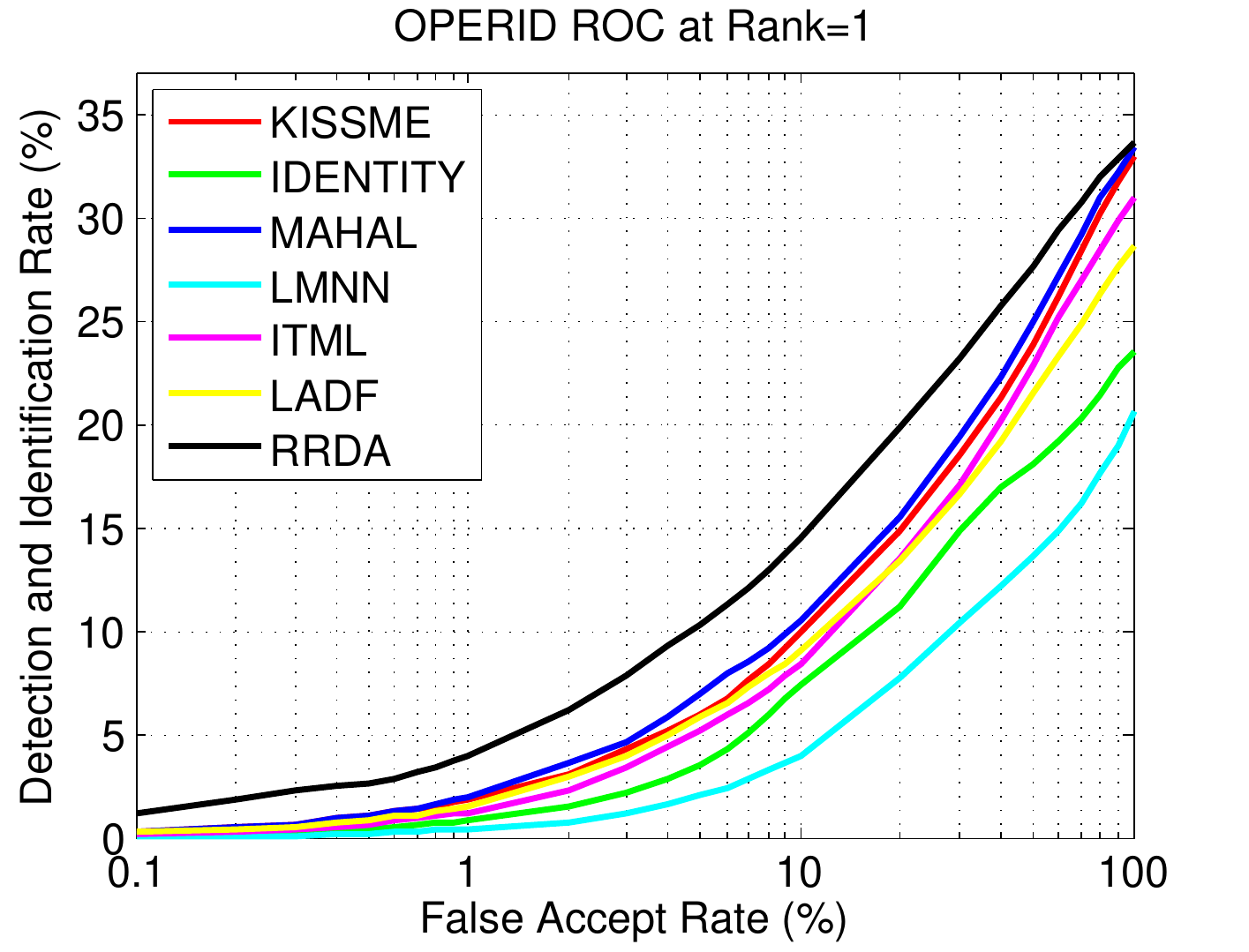}
\caption{ROC curves with rank=1.}
\label{fig:rank1}
\end{figure}

\begin{figure}
\centering
\includegraphics[width=80mm]{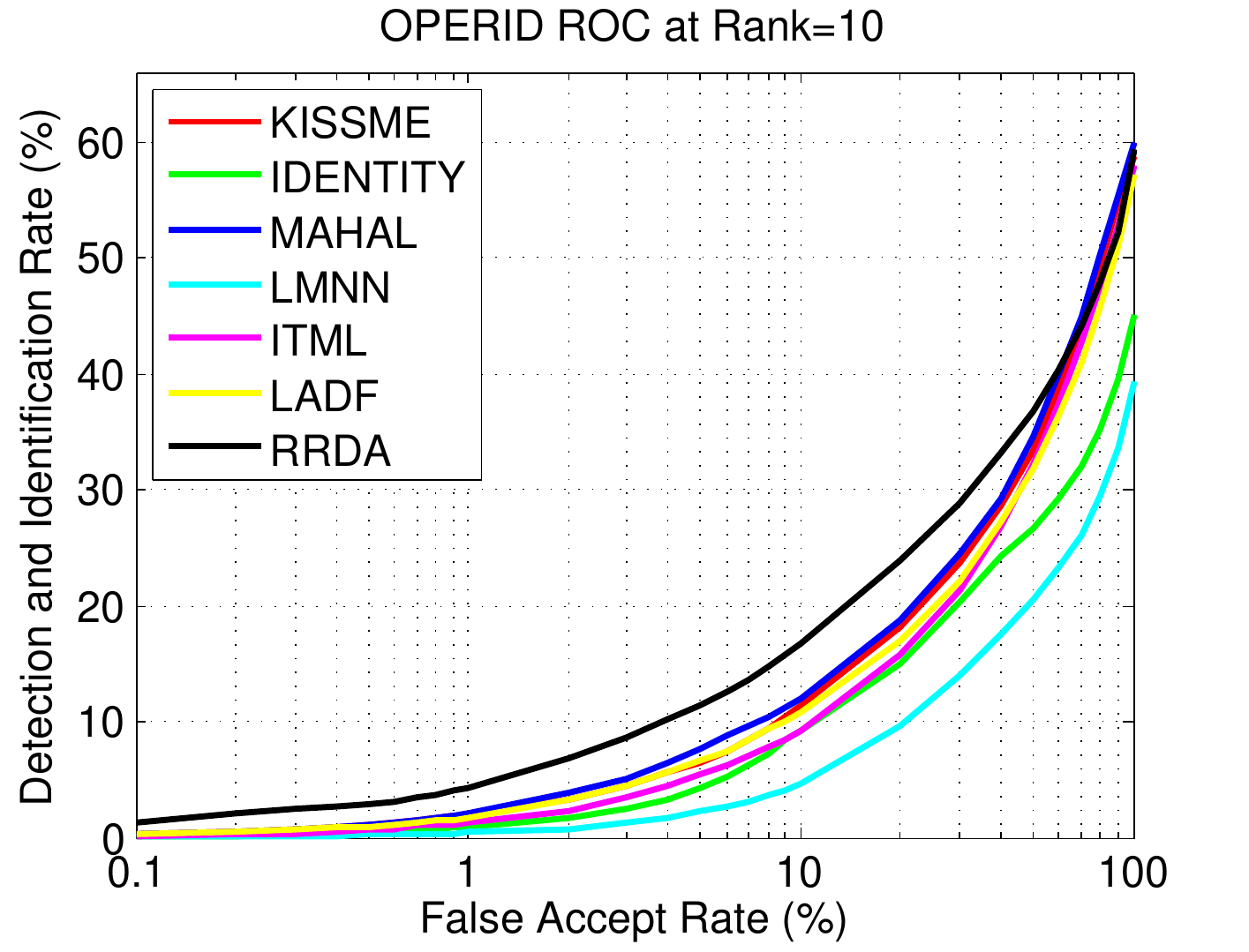}
\caption{ROC curves with rank=10.}
\label{fig:rank10}
\end{figure}

In this section, we summarize the baseline performance of the open-set person re-identification problem, evaluated on our collected multi-camera database. We follow the procedure described in the benchmark protocol, resulting in the ROC curves with rank=1 and rank=10 shown in Figs. \ref{fig:rank1} and \ref{fig:rank10}. Recall that FAR=100\% corresponding to the traditional closed-set person re-identification problem. From the figures it can be seen that the closed-set person re-identification performances of the evaluated algorithms are promising, compared to the performance observed from other databases. However, it is clear that the open-set person re-identification performance significantly drops with the decreasing value of FAR. This means that the current algorithm is still far from satisfactory in the real application aspect of view. Nevertheless, the simple RRDA algorithm is shown to be quite superior for open-set person re-identification, compared to other evaluated algorithms.

\begin{figure}
\centering
\includegraphics[width=80mm]{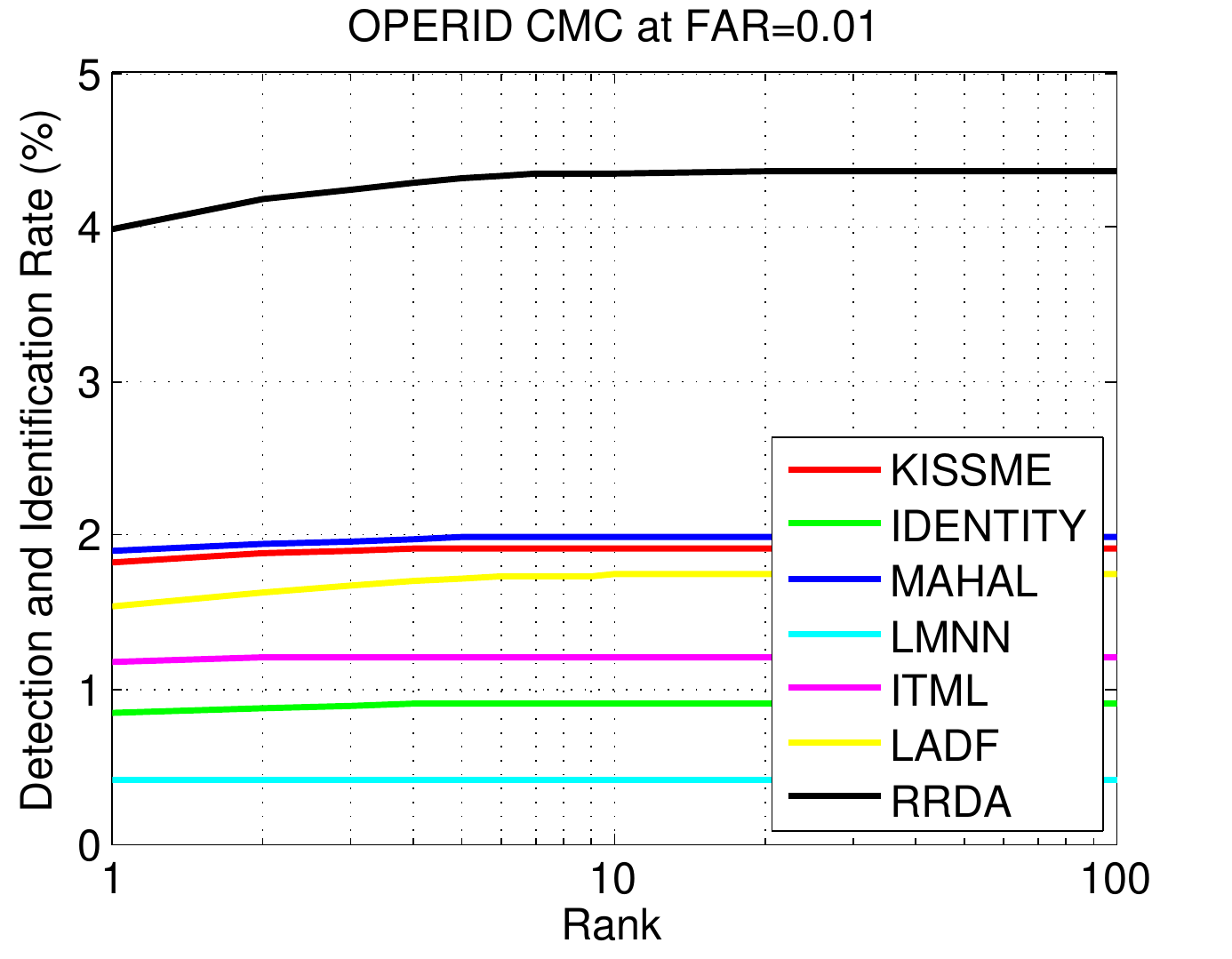}
\caption{CMC curves with FAR=1\%.}
\label{fig:far1}
\end{figure}

\begin{figure}
\centering
\includegraphics[width=80mm]{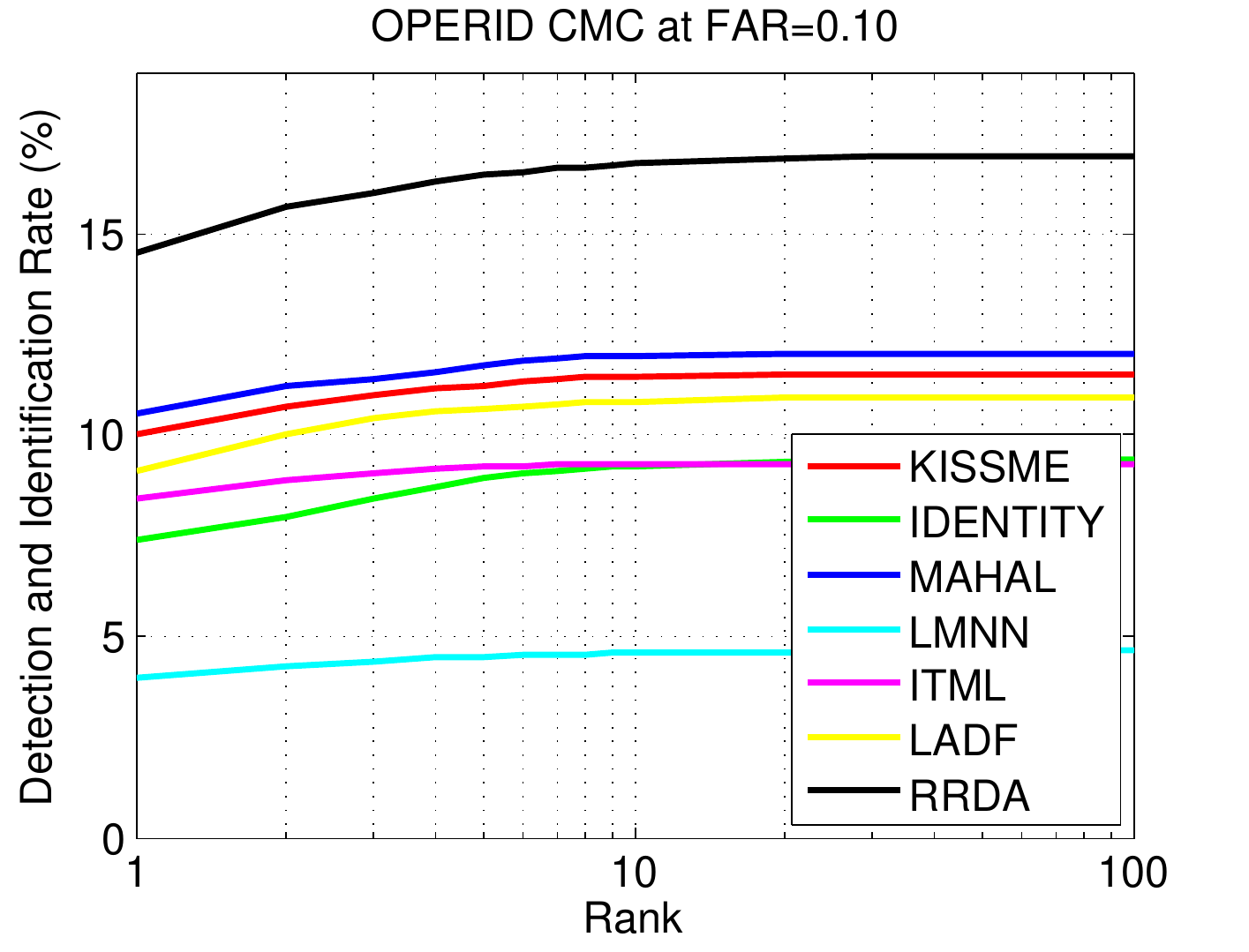}
\caption{CMC curves with FAR=10\%.}
\label{fig:far10}
\end{figure}

We also calculated the performance at fixed FARs, and generated CMC curves under FAR=1\% and FAR 10\%, as shown in Fig. \ref{fig:far1} and Fig. \ref{fig:far10}. From these curves it is observed that all algorithms perform very poor at the two FAR points. What is worse, it can also be observed that increasing the number of retrievals after rank 10 helps very little in improving the performance. This is because genuine matching scores at lower ranks are hardly be larger than the decision threshold.

To be specific, we summarize some results in Table \ref{tab:results}. The results clearly indicate a demand for improving the open-set person re-identification performance, since all algorithms did not exceed 17\%. Nevertheless, it can be observed that RRDA is the best performer among the evaluated algorithms. In summary, the open-set person re-identification problem is still far from solved, and deserve further research and development.

\begin{table}
\centering
\caption{Detection and identification rates (\%) of metric learning algorithms under several operating points.}\label{tab:results}
\begin{tabular}{|c|c|c|c|c|}
\hline
	& \multicolumn{2}{|c}{FAR=1\%} &\multicolumn{2}{|c|}{FAR=10\%}\\
\cline{2-5}
	&Rank=1 &Rank=10	&Rank=1 &Rank=10	\\
\hline
KISSME	 &1.82  &1.92      &9.99  &11.46    \\
LMNN	 &0.41  &0.41      &3.97  &4.58    \\
ITML	 &1.18  &1.21      &8.39  &9.27    \\
MAHAL	 &1.89	&1.99      &10.50 &11.97   \\
IDENTITY &0.84	&0.91      &7.36  &9.21    \\
LADF     &1.53  &1.74      &9.11  &10.82    \\
RRDA	 &\textbf{3.99}	&\textbf{4.35}	   &\textbf{14.51}  &\textbf{16.72}  \\
\hline
\end{tabular}
\end{table}

\section{Conclusions}\label{sec:summary}
In this paper, we have considered the open-set person re-identification problem, which includes two sub-tasks, detection and identification. We have contributed a database collected from a video surveillance setting of 6 cameras, with 200 persons and 7,413 images segmented. Based on this database, we have developed a benchmark protocol for evaluating the performance under the open-set person re-identification scenario. Several popular metric learning algorithms for person re-identification have been evaluated as baselines. From the baseline performance, we conclude that the open-set person re-identification problem is still largely unresolved, thus further attention and effort is needed. We have released the collected database OPeRID v1.0, as well as the extracted features and the benchmark tool to advance research along this direction.

\bibliographystyle{splncs}
\bibliography{../Bib/Liao}
\end{document}